\documentclass[runningheads,a4paper]{llncs}

\usepackage{amssymb}
\usepackage{graphicx}

\usepackage{amsmath}
\usepackage[lofdepth,lotdepth]{subfig}

\usepackage{tikz}
\usetikzlibrary{arrows}

\usepackage{url}
\urldef{\maildavid}\path|david.robben@kuleuven.be|

\begin{document}

\mainmatter  

\title{Perfusion parameter estimation\\using neural networks and data augmentation}
\titlerunning{Perfusion parameter estimation using neural networks}
\toctitle{Perfusion parameter estimation using neural networks and data augmentation}

\tocauthor{David Robben}

\author{David Robben \and Paul Suetens }
\authorrunning{Robben and Suetens}

\institute{
Medical Image Computing (ESAT/PSI), KU Leuven, Belgium
}

\maketitle

\begin{abstract}	
Perfusion imaging plays a crucial role in acute stroke diagnosis and treatment decision making.
Current perfusion analysis relies on deconvolution of the measured signals, an operation that is mathematically ill-conditioned and requires strong regularization.
We propose a neural network and a data augmentation approach to predict perfusion parameters directly from the native measurements.
A comparison on simulated CT Perfusion data shows that the neural network provides better estimations for both CBF and Tmax than a state of the art deconvolution method,
 and this over a wide range of noise levels.
The proposed data augmentation enables to achieve these results with less than 100 datasets.

\end{abstract}


\section{Introduction}
In stroke perfusion imaging \cite{Fieselmann2011} a series of 3D MR or CT images of the brain are acquired after injection of a contrast bolus. These images show the contrast agent -- and hence the blood -- flow in and out of the brain. As such, we have in each voxel a time series that shows the change in image intensity due the contrast agent. This intensity change can be converted to the concentration of contrast agent. This imaging modality plays a crucial role in stroke diagnosis, allowing to measure perfusion parameters such as cerebral blood flow (CBF), blood volume (CBV) and arrival time in each voxel of the brain. These perfusion parameters allow to assess to what extent the brain is affected by the stroke, distinguishing the core of the infarct (dead tissue), the penumbra (tissue at risk) and the healthy tissue. The volumes of the core and penumbra are essential to decide on the treatment plan of an acute stroke patient~\cite{AlbersDefuse3}.

The native images acquired in perfusion imaging, be it CT perfusion or MR perfusion, are not directly interpretable. Typically, a deconvolution analysis is performed, where the concentration time series that are measured in each cerebral voxel (the time concentration curve or TCC), are deconvolved with the so-called arterial input function (AIF), which is the concentration time series measured in one of the large feeding arteries of the brain. The deconvolved time series are no longer influenced by the contrast injection protocol or the cardiac status of the subject. They are impulse response functions (IRF), the signal that theoretically would be measured if the injection of contrast agent where a Dirac impulse directly into the feeding artery of the brain. It has been shown that, under reasonable assumptions, the maximum of this deconvolved time series is proportional to the cerebral blood flow (CBF). This parameter is strongly predictive for the health of the tissue and is currently used in clinical practice to identify the infarct core: voxels with a CBF less than 30\% compared to the other side of the brain are considered to be core \cite{Albers2016}. The time when the deconvolved time series reaches its maximum, is called Tmax, and increased values are indicative for tissue is at risk. A threshold of 6 seconds is currently clinically used to determine the perfusion lesion (i.e. the combined core and penumbra) of the infarct \cite{Albers2016}.

Indeed, deconvolution plays a central role in perfusion analysis. However, deconvolution is a mathematically ill-posed problem, and given the relatively low signal to noise ratio of perfusion images, a successful implementation of perfusion analysis requires duly attention to this problem. First, there is need for proper preprocessing: motion correction, temporal and spatial smoothing, and possibly spatial downsampling. Second, the deconvolution is regularized, suppressing the high frequency signal in the reconstructed impulse response function. This is typically done in singular value decomposition (SVD) based deconvolution by regularizing the singular values, e.g. using Tikhonov regularization.
Nevertheless, the deconvolution-based perfusion parameters remain noise sensitive and research for improved algorithms \cite{Boutelier2012} or even deconvolution-free summary parameters \cite{Meijs2016} remains ongoing.

Recently, several works have proposed to use machine learning techniques to estimate, based on the perfusion and treatment parameters, how the final infarct will look \cite{Wu2001,Kemmling2015,Maier2016,Nielsen2018}.
However, all these approaches first perform a deconvolution analysis -- which suffers from the earlier mentioned problems --
 and then use the perfusion parameters as input features for the machine learning algorithms. 
One notable exception is \cite{Pinto2018} who use both the perfusion parameters and the native measurements as input to their method.

In this work, we show on simulated data that a neural network can learn to perform this deconvolution and achieve more accurate estimations of CBF and Tmax
 than a state of the art deconvolution technique can.
Additionally, we show how a perfusion specific data augmentation can be used to learn this deconvolution from a relatively small number of training samples.
Knowing that a neural network is able to learn how to perform the deconvolution, opens possibilities for new research,
 where the final infarct is predicted directly from the native images, bypassing the standard deconvolution as a preprocessing step.


\section{Methods}
We will compare a state of the art deconvolution approach with our proposed neural network-based approach.

\subsection{Baseline: Tikhonov regularized SVD-based deconvolution}
As a baseline, we use Tikhonov regularized SVD-based deconvolution \cite{Fieselmann2011}
 with the Volterra discretization scheme \cite{Sourbron2007} -- to which we will simply refer as deconvolution.
This method takes the AIF and TCC and will produce the IRF.
From the IRF, we can infer both the CBF and the Tmax:
\begin{flalign}
\text{CBF}		&= \frac{1}{\rho} \max_t \text{IRF}(t)
\\
\text{Tmax} 	&= \operatorname*{argmax}_t \text{IRF}(t)
.
\end{flalign}
The IRF has as unit $1/s$ and hence the CBF $ml/g/s$ (being $ml$ blood per $g$ of brain tissue per $s$).
When reporting CBF values, we will follow the common practice of using $ml/100g/min$ .
The IRF is discretized, so to produce continuous estimates of Tmax, the IRF is fitted with a quadratic spline.

The deconvolution has one hyperparameter, the relative regularization parameter $\lambda_{rel}$ which sets the amount of filtering.
For each experiment, the optimal regularization strength is found by evaluating the performance on the training set for a range of possible values ($0.01 * 2^{0,1,2,..,9}$).

\subsection{Proposed neural network}
Through experimentation we found that even simple networks succeed at this task and the following network is used for all experiments.
Our network has two input layers, one for the AIF, and one for the TCC. 
The two input layers are concatenated, followed by two fully connected layers with each 30 neurons and end in a single output neuron.
The fully connected layers have a PreLU activation and the output layer has no activation function.
The output is a single value, depending on the experiment, either the CBF or the Tmax.

Note that while it is possible to provide spatial context to the network -- e.g. the TCC of the neighboring voxels, which would most certainly improve the performance -- this is out of scope for this work and would make the comparison between the two approaches unfair.

The network is trained in a supervised fashion by feeding examples of an AIF and a TCC with known CBF or Tmax
 and minimizing the absolute difference between prediction and ground truth.
The optimization is performed by stochastic gradient descent with learning rate 0.01, Nesterov momentum of 0.9 and mini-batches of 2048 samples.

\subsection{Proposed data augmentation}
To limit the amount of training data that is required to train this network, we propose a perfusion specific data augmentation. 
The key insight is that the perfusion measurements are a linear time invariant system.
This means that, if the contrast injection was a bit later, both the AIF and the TCC would show the same delay.
Similarly, if the injection was earlier, all curves should shift to the left.
However, the IRF (and hence the Tmax and CBF) would remain the same in both cases.
If the concentration of the iodine or gadolinium in the contrast agent were a fraction higher or lower, the concentration curves should be changed with the same fraction.
But again the IRF would remain untouched.

Hence, we will create additional samples from a single training sample by applying a random time shift (earlier or later) and a random scaling to both AIF and TCC.
In our experiments, the shift is randomly chosen between -1 and 2 time points (since our measurments are discrete) and the scaling has a uniform distribution between [0.7,1.3].
This augmentation enhances the number of different training samples the network sees, reducing the required size of our actual training set.


\section{Experiments}
To compare the deconvolution with the proposed methods, we need to compare the predictions with the ground truth.
However, ground truth measures of the CBF or Tmax are not possible in vivo.
Hence we perform a series of experiments on simulated data.

\subsection{Data}
In the simulations, we model the arterial input function (AIF) and the impuls response function (IRF) as gamma variates:
\begin{flalign}
\gamma(t ; t_0, \alpha, \beta, A)
=
A * (t-t_0)^\alpha * e^{- \frac{t-t_0}{\beta} }
.
\end{flalign}
The different parameters are chosen randomly such that the resulting curves resemble the curves measured with CTP.
The values $t_0, \alpha, \beta$ are chosen from uniform distributions.
For the AIF:  
$t_0 \in [0,15]$,
$\alpha \in [1.5,3.5]$,
$\beta \in [1.5,3.5]$.
For the IRF: 
$t_0 \in [0,10]$, 
$\alpha \in [0,0.5]$,
$\beta \in [2.5,4.5]$.
    
For the AIF, the value of $A$ is chosen such that the AIF's maximum is uniformly distributed between 100 and 500 HU.
\footnote{In CTP, the change in intensity is proportional to the contrast agent concentration and hence deconvolution techniques yield the same results whether applied on the intensity changes or on the concentration. The former have the advantage of being more easily interpretable, and hence we will express the AIF and TCC in Hounsfield Units (HU).}
For the IRF, the value of $A$ is chosen such that the integral of the IRF (i.e. the CBV) is uniformly distributed between 0.1\% and 6\%. 
For the experiments about Tmax estimation, a different range is chosen (between 2\% and 6\%)
 since the deconvolution-based Tmax estimation performs poorly on weak signals.
Keeping the full range would move the comparison in favor for the proposed method while not being relevant in clinical practice. 

\begin{figure}[hbt]
	\centering
	\subfloat[An example with $\sigma=1$.]{
		\includegraphics[width=.4\textwidth]{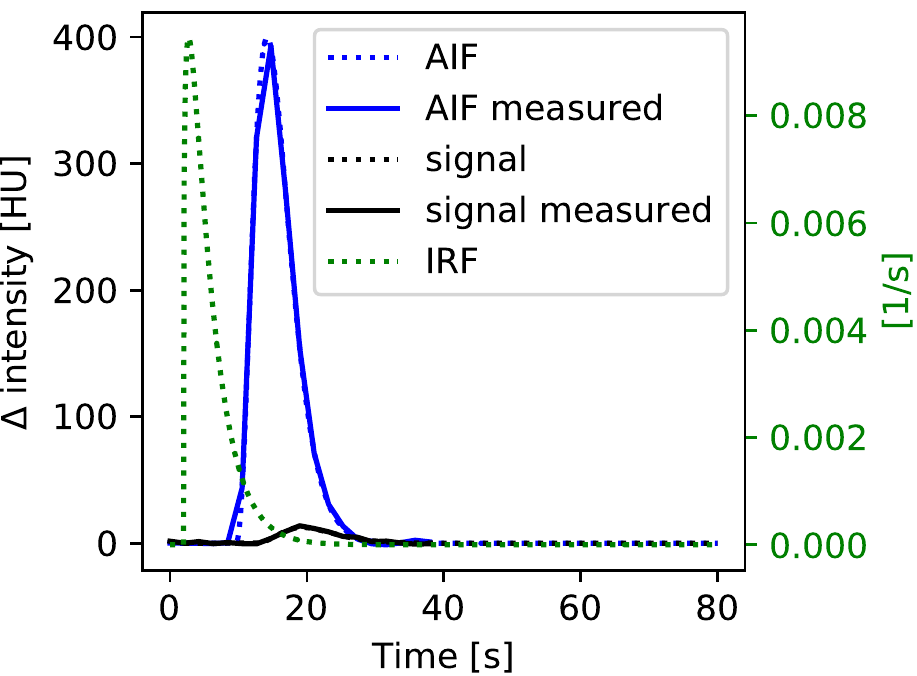}
		\label{fig:simulation_example}
	}
	\subfloat[Histograms of the perfusion parameters.]{
		\includegraphics[width=.6\textwidth]{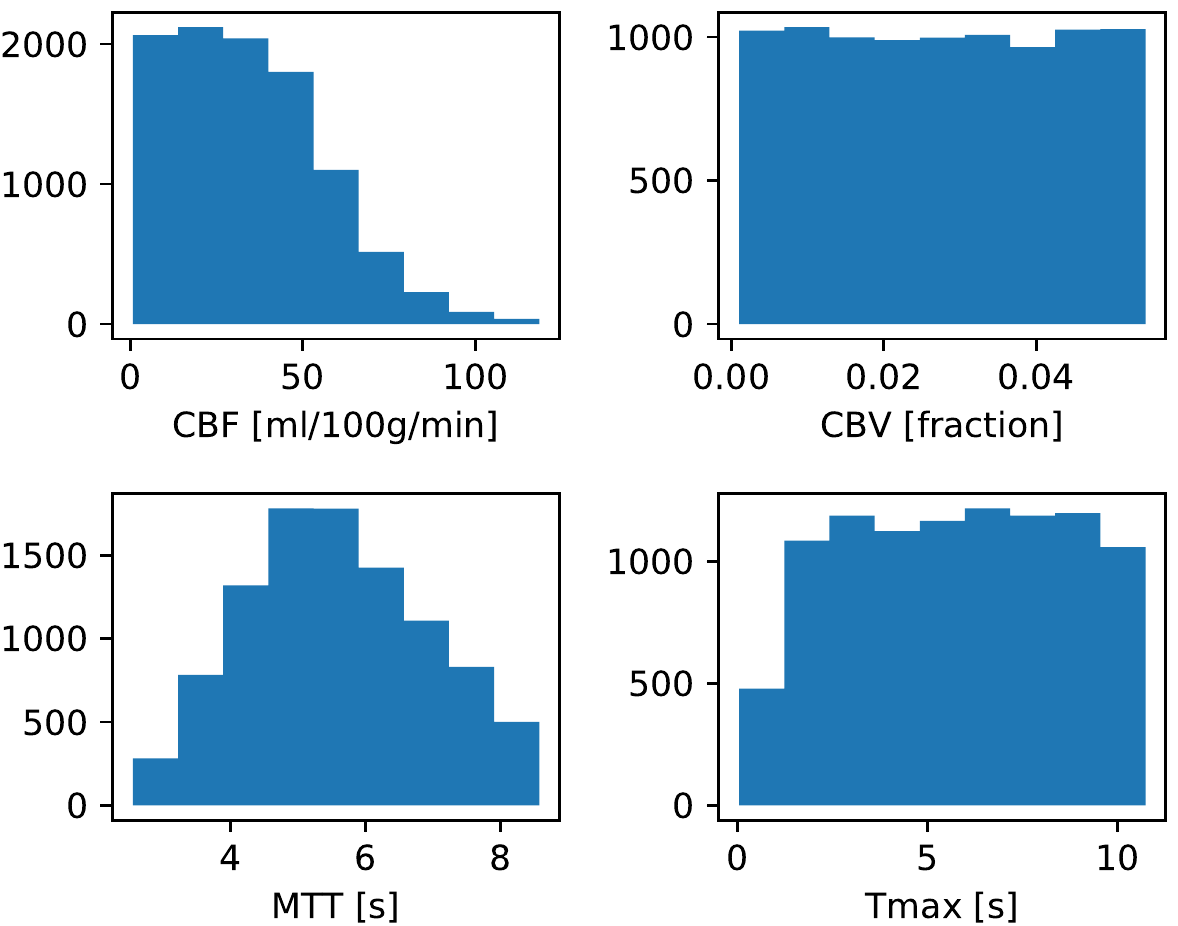}
		\label{fig:simulation_distributions}
	}
	\caption{The simulated data. }
	\label{fig:simulation}
\end{figure}

The TCC is obtained by convolving the simulated AIF and IRF.
Finally, the AIF and TCC are sampled (19 samples over a time span of 40 seconds) and Gaussian noise is added ($\sigma$ varies from 0.1 to 3.2, and is given later for each specific experiment).
Figure~\ref{fig:simulation} shows a randomly simulated sample and the resulting distribution of CBF, CBV, MTT and Tmax.

\subsection{SVD-based deconvolution versus proposed neural network}
We aim to compare how well the deconvolution approach and the neural network are able to estimate the CBF and Tmax from the AIF and TCC.
The performance of the Tmax estimation is measured using the mean absolute difference (MAD) between the true and estimated value.
For the CBF, we also use the MAD, but only after scaling all the estimates with an optimal scaling factor (i.e. the one that minimizes the MAD).
This is warranted since in clinical practice the relative CBF is used to predict the infarct core.

\begin{figure}[hbt]
	\centering
	\subfloat[Mean absolute difference between the true and estimated CBF (after scaling) in the test set for various noise levels.]{
		\includegraphics[width=.5\textwidth]{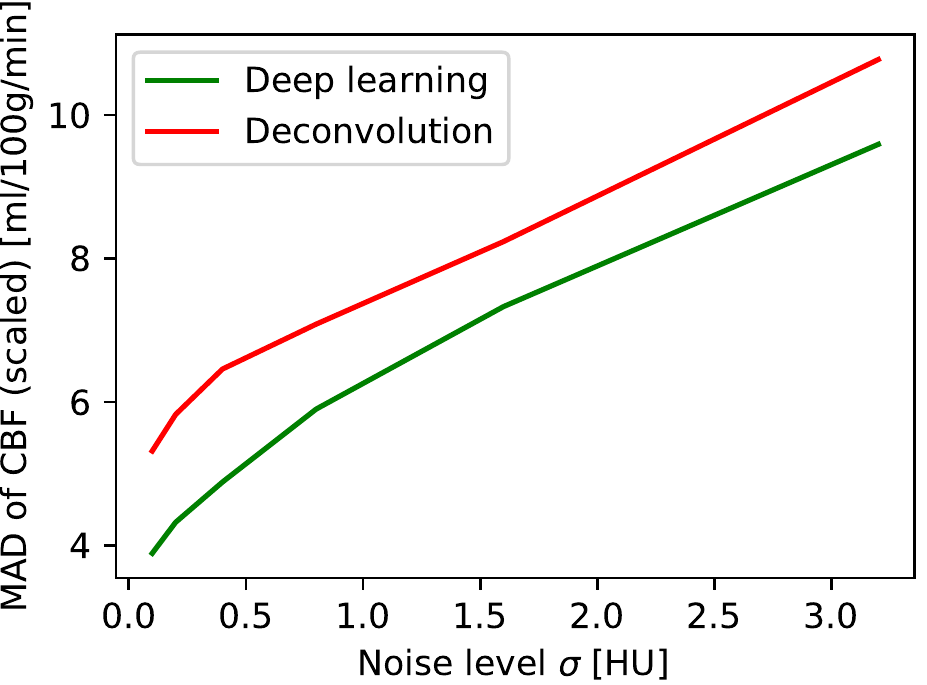}
		\label{fig:dl_vs_deconv_cbf}
	}
	\subfloat[Scatter plot of test samples at $\sigma=1$.]{
		\includegraphics[width=.5\textwidth]{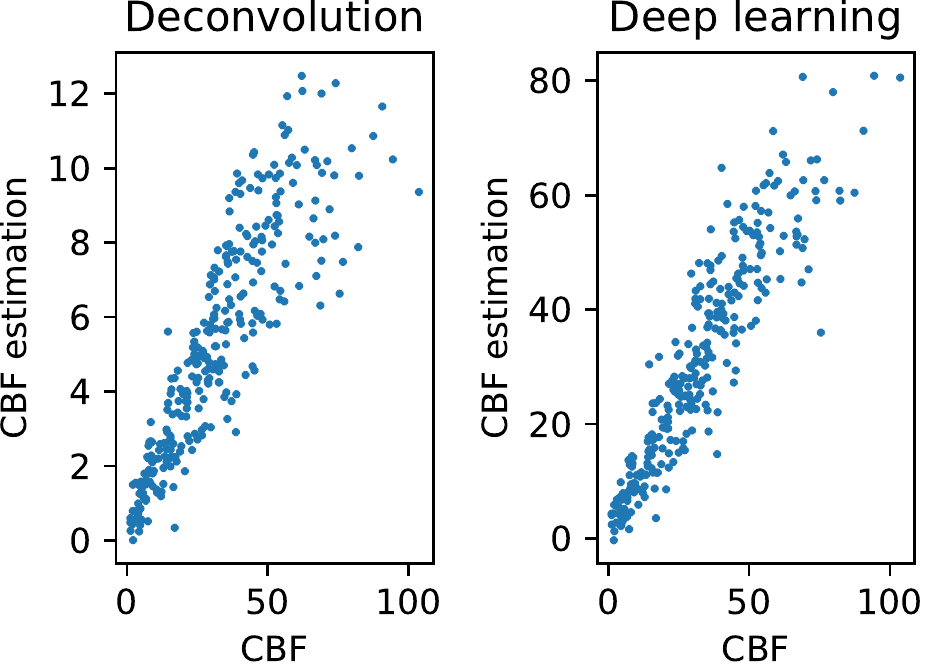}
		\label{fig:dl_vs_deconv_cbf_example}
	}
	\\
	\subfloat[Mean absolute difference between the true and estimated Tmax in the test set for various noise levels.]{
		\includegraphics[width=.5\textwidth]{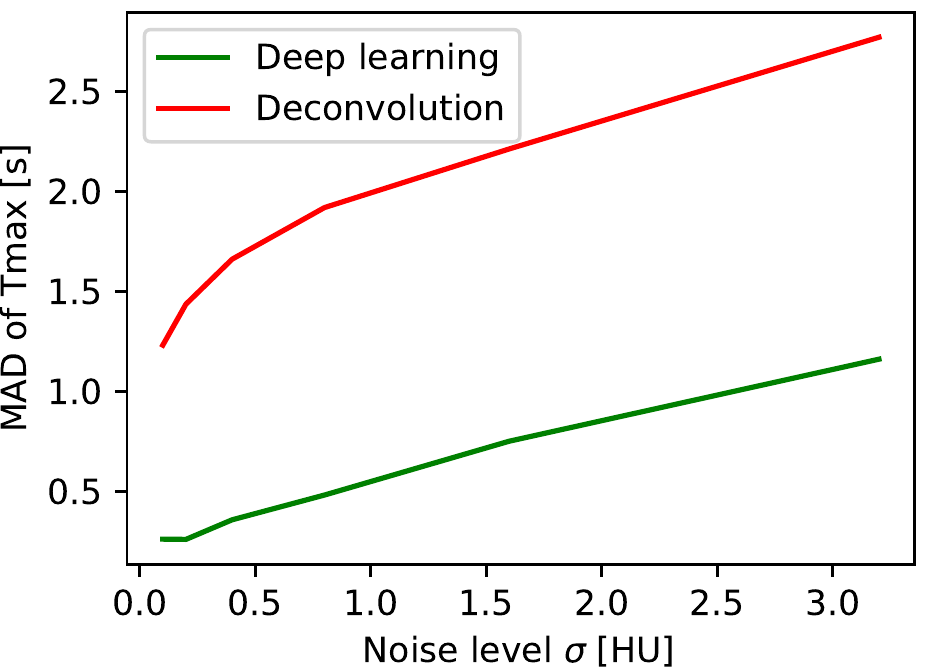} 
		\label{fig:dl_vs_deconv_tmax}
	}
	\subfloat[Scatter plot of test samples at $\sigma=1$.]{
		\includegraphics[width=.5\textwidth]{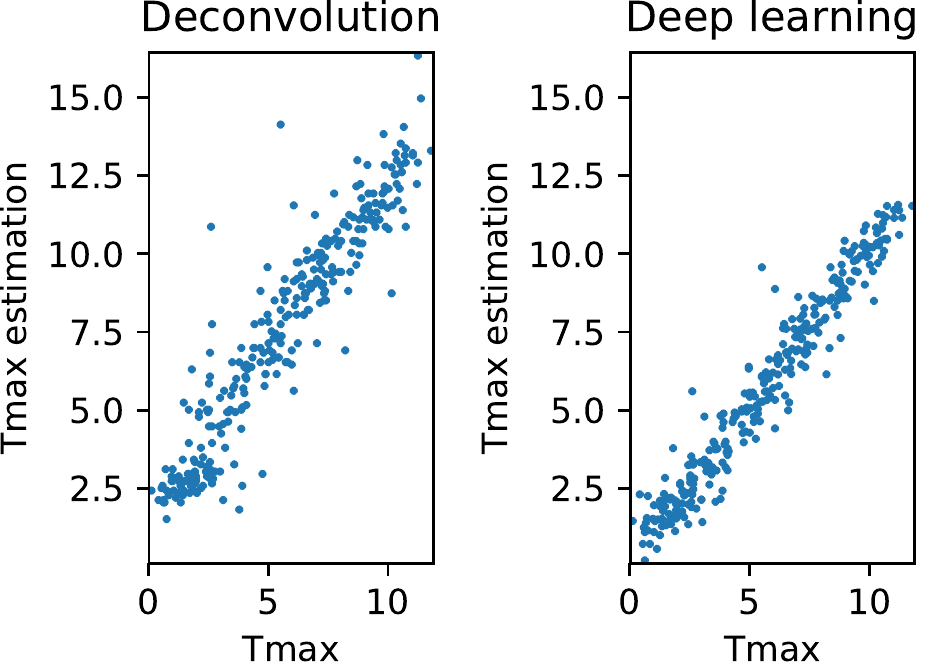}
		\label{fig:dl_vs_deconv_tmax_example}
	}
	\caption{Comparison between estimations produced by the neural network and the SVD-based deconvolution.}
	\label{fig:dl_vs_deconv}
\end{figure}

For a range of noise values, we generate a training set (1M samples) and a testing set (10k samples),
 where a sample consists the AIF, the TCC and the ground truth perfusion parameter (CBF or Tmax). 
For each noise level, the optimal amount of Tikhonov regularization is determined on the training set and that value is used on the test set.
The neural network is trained for one epoch on all training samples and subsequently predicts the test set.
Fig.~\ref{fig:dl_vs_deconv_cbf} and~\ref{fig:dl_vs_deconv_tmax} summarize the results and show an improvement for both measures on all noises levels.
Fig.~\ref{fig:dl_vs_deconv_cbf_example} and~\ref{fig:dl_vs_deconv_tmax_example} show the distribution of the predictions at a single noise level.


\subsection{Data augmentation}
\begin{figure}[hbt]
	\centering
		\subfloat{
		\includegraphics[width=.5\textwidth]{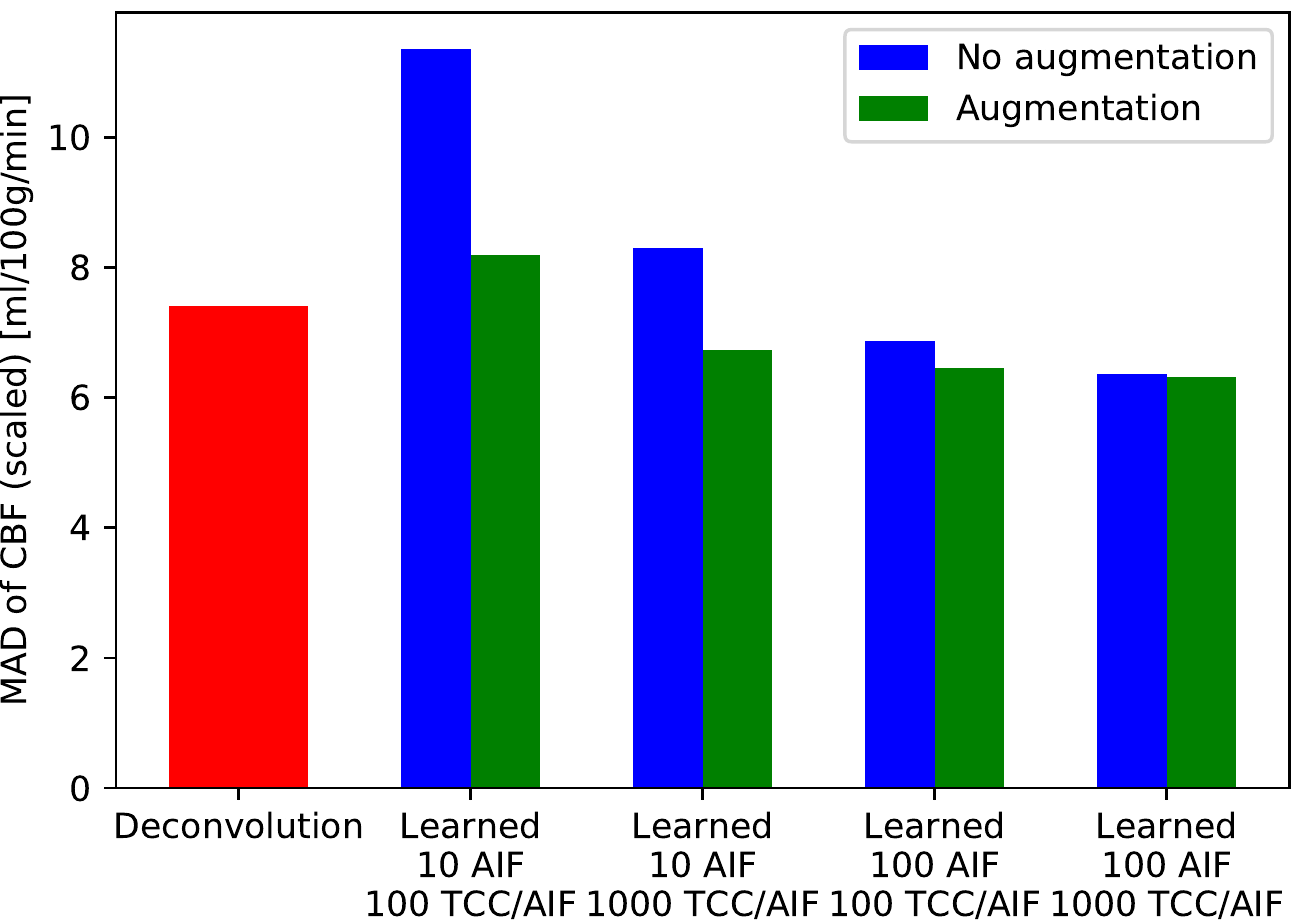}
		\label{fig:data_augmentation_cbf}
	}
	\subfloat{
		\includegraphics[width=.5\textwidth]{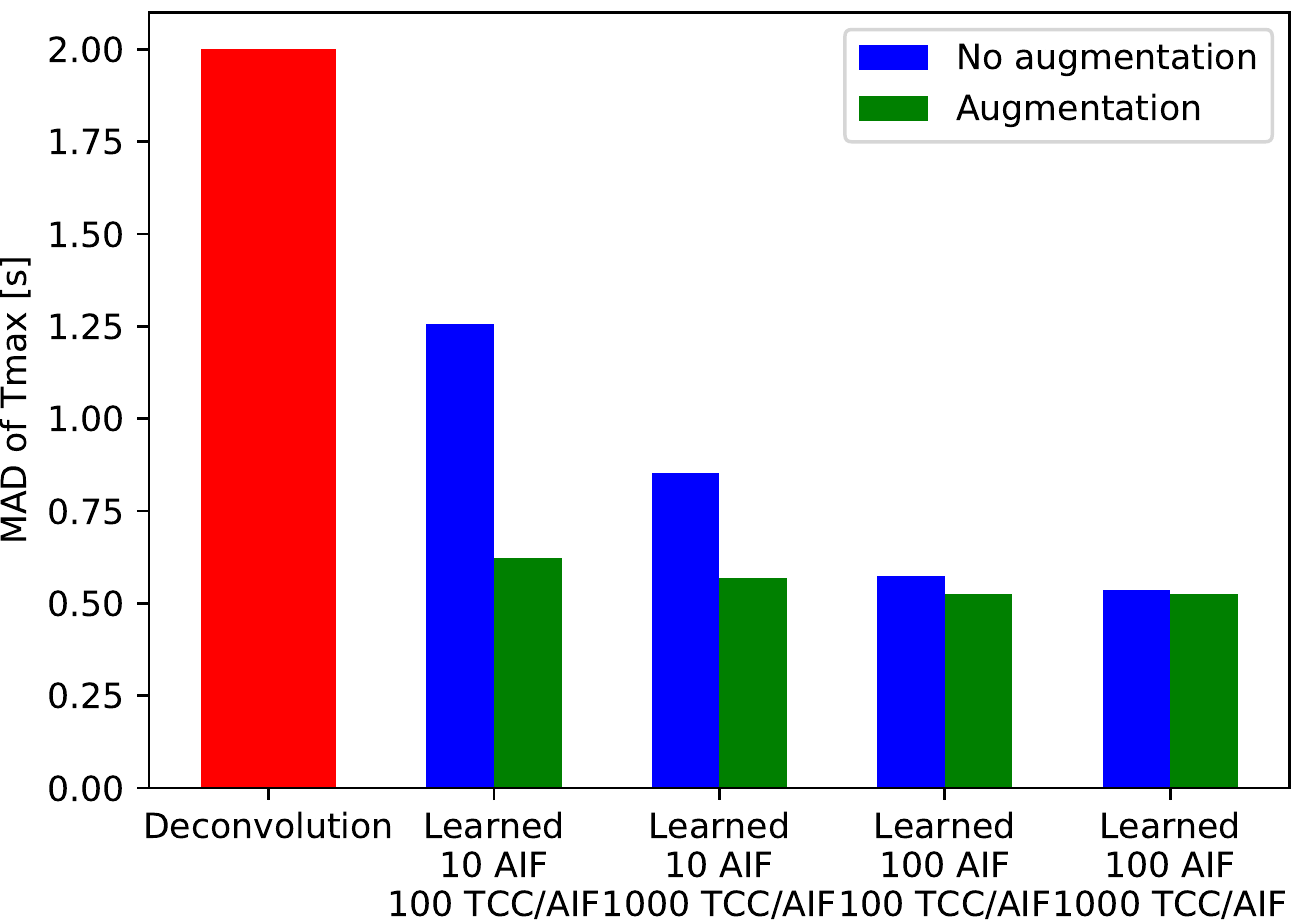}
		\label{fig:data_augmentation_tmax}
	}
	\caption{Influence of the number of training samples and data augmentation on the performance of the neural network.
	}
	\label{fig:data_augmentation}
\end{figure}

In the previous experiment, we trained on 1 million samples, each with a different AIF and hence each corresponding to a different acquisition.
Such large training sets are not realistic,
 and in this experiment we explore how many samples are necessary and whether our proposed data augmentation can lower that number.
Using a fixed noise level of $\sigma=1$, we create training sets of various sizes,
 by varying the number of AIFs (which corresponds the number of acquisitions) and the number of TCCs per AIF.
The proposed data augmentation is used to increase the number of samples with a factor 10.
We use the same performance metrics and training method as in the previous section,
 with the number of training epochs adapted such that each network is trained for the same number of iterations (corresponding to one epoch with 1M samples).
Fig.~\ref{fig:data_augmentation} summarizes the results, showing that less than 100 acquisitions are sufficient and that the proposed data augmentation can lead to large improvements, especially in situations with limited training data.

\section{Discussion and conclusion}
We proposed a simple neural network and a data augmentation approach to predict perfusion parameters from native perfusion measurements (i.e. the AIF and TCC).
A comparison on simulated data shows that the neural network provides better estimations for both CBF and Tmax than a state of the art deconvolution method, and this over a wide range of noise levels.
Using the proposed data augmentation, it is feasible to achieve these results with less than 100 datasets.

Earlier, Ho et al. \cite{Ho2016} showed that a neural network can learn how to deconvolve.
They trained and tested a CNN on MR perfusion datasets while using as a ground truth the perfusion parameters obtained by an SVD-based deconvolution method.
They showed that a neural network can produce a reasonable approximation for the various perfusion parameters.
In this work, we go one step further, and show that a neural network can outperform deconvolution methods on simulated CT Perfusion data. 

We see two main directions for future research.
First, one could investigate how neural network based perfusion parameter estimation works on real data.
As mentioned earlier, validation of such an approach is difficult since ground truth measurements are not available.
However, it might be possible to produce convincing evidence by comparing different modalities from the same subject
 or by comparing results on high- and low-resolution versions of the same dataset.
Second, one could revisit the earlier mentioned methods that attempt to estimate the final infarct from perfusion parameter maps.
Especially for neural network based approaches, such as \cite{Nielsen2018},
 it might be beneficial to provide the original time series to the network instead of the perfusion parameters.
Having the network perform the deconvolution operation implicitly might lead to better estimates of the CBF and Tmax, 
 or -- even more promising -- might lead the network to learn new perfusion parameters that are even more predictive.
 

\paragraph*{Acknowledgement} David Robben is supported by an innovation mandate of Flanders Innovation \& Entrepreneurship (VLAIO). 

\bibliographystyle{splncs03}
\bibliography{references_without_url}

\begin{thebibliography}{10}
\providecommand{\url}[1]{\texttt{#1}}
\providecommand{\urlprefix}{URL }

\bibitem{Albers2016}
Albers, G.W., Goyal, M., Jahan, R., Bonafe, A., Diener, H.C., Levy, E.I.,
  Pereira, V.M., Cognard, C., Cohen, D.J., Hacke, W., Jansen, O., Jovin, T.G.,
  Mattle, H.P., Nogueira, R.G., Siddiqui, A.H., Yavagal, D.R., Baxter, B.W.,
  Devlin, T.G., Lopes, D.K., Reddy, V.K., {De Rochemont}, R.D.M., Singer, O.C.,
  Bammer, R., Saver, J.L.: {Ischemic core and hypoperfusion volumes predict
  infarct size in SWIFT PRIME}. Annals of Neurology  79(1),  76--89 (2016)

\bibitem{AlbersDefuse3}
Albers, G.W., Marks, M.P., Kemp, S., Christensen, S., Tsai, J.P.,
  Ortega-Gutierrez, S., McTaggart, R.A., Torbey, M.T., Kim-Tenser, M.,
  Leslie-Mazwi, T., Sarraj, A., Kasner, S.E., Ansari, S.A., Yeatts, S.D.,
  Hamilton, S., Mlynash, M., Heit, J.J., Zaharchuk, G., Kim, S., Carrozzella,
  J., Palesch, Y.Y., Demchuk, A.M., Bammer, R., Lavori, P.W., Broderick, J.P.,
  Lansberg, M.G.: {Thrombectomy for Stroke at 6 to 16 Hours with Selection by
  Perfusion Imaging}. New England Journal of Medicine  378(8),  708--718 (feb
  2018)

\bibitem{Boutelier2012}
Boutelier, T., Kudo, K., Pautot, F., Sasaki, M.: {Bayesian hemodynamic
  parameter estimation by bolus tracking perfusion weighted imaging}. IEEE
  Transactions on Medical Imaging  31(7),  1381--1395 (2012)

\bibitem{Fieselmann2011}
Fieselmann, A., Kowarschik, M., Ganguly, A., Hornegger, J., Fahrig, R.:
  {Deconvolution-Based CT and MR Brain Perfusion Measurement: Theoretical Model
  Revisited and Practical Implementation Details}. International Journal of
  Biomedical Imaging  2011,  1--20 (2011)

\bibitem{Ho2016}
Ho, K.C., Scalzo, F., Sarma, K.V., El-Saden, S., Arnold, C.W.: {A temporal deep
  learning approach for MR perfusion parameter estimation in stroke}. In: 2016
  23rd International Conference on Pattern Recognition (ICPR). pp. 1315--1320.
  IEEE (dec 2016)

\bibitem{Kemmling2015}
Kemmling, A., Flottmann, F., Forkert, N.D., Minnerup, J., Heindel, W.,
  Thomalla, G., Eckert, B., Knauth, M., Psychogios, M., Langner, S., Fiehler,
  J.: {Multivariate Dynamic Prediction of Ischemic Infarction and Tissue
  Salvage as a Function of Time and Degree of Recanalization}. Journal of
  Cerebral Blood Flow {\&} Metabolism  35(9),  1397--1405 (sep 2015)

\bibitem{Maier2016}
Maier, O., Menze, B.H., von~der Gablentz, J., H{\"{a}}ni, L., Heinrich, M.P.,
  Liebrand, M., Winzeck, S., Basit, A., Bentley, P., Chen, L., Christiaens, D.,
  Dutil, F., Egger, K., Feng, C., Glocker, B., G{\"{o}}tz, M., Haeck, T.,
  Halme, H.L., Havaei, M., Iftekharuddin, K.M., Jodoin, P.M., Kamnitsas, K.,
  Kellner, E., Korvenoja, A., Larochelle, H., Ledig, C., Lee, J.H., Maes, F.,
  Mahmood, Q., Maier-Hein, K.H., McKinley, R., Muschelli, J., Pal, C., Pei, L.,
  Rangarajan, J.R., Reza, S.M.S., Robben, D., Rueckert, D., Salli, E., Suetens,
  P., Wang, C.W., Wilms, M., Kirschke, J.S., Kr{\"{a}}mer, U.M., M{\"{u}}nte,
  T.F., Schramm, P., Wiest, R., Handels, H., Reyes, M.: {ISLES 2015 - A public
  evaluation benchmark for ischemic stroke lesion segmentation from
  multispectral MRI}. Medical Image Analysis  35,  250--269 (jan 2017)

\bibitem{Meijs2016}
Meijs, M., Christensen, S., Lansberg, M.G., Albers, G.W., Calamante, F.:
  {Analysis of perfusion MRI in stroke: To deconvolve, or not to deconvolve}.
  Magnetic Resonance in Medicine  76(4),  1282--1290 (2016)

\bibitem{Nielsen2018}
Nielsen, A., Hansen, M.B., Tietze, A., Mouridsen, K.: {Prediction of Tissue
  Outcome and Assessment of Treatment Effect in Acute Ischemic Stroke Using
  Deep Learning}. Stroke  49(6),  1394--1401 (jun 2018)

\bibitem{Pinto2018}
Pinto, A., Pereira, S., Meier, R., Alves, V., Wiest, R., Silva, C.A., Reyes,
  M.: {Enhancing Clinical MRI Perfusion Maps with Data-Driven Maps of
  Complementary Nature for Lesion Outcome Prediction}. In: Frangi, A.F.,
  Schnabel, J.A., Davatzikos, C., Alberola-L{\'{o}}pez, C., Fichtinger, G.
  (eds.) MICCAI. Lecture Notes in Computer Science, vol. 11072, pp. 107--115.
  Springer International Publishing (2018)

\bibitem{Sourbron2007}
Sourbron, S., Luypaert, R., Morhard, D., Seelos, K., Reiser, M., Peller, M.:
  {Deconvolution of bolus-tracking data: a comparison of discretization
  methods}. Physics in Medicine and Biology  52(22),  6761--6778 (nov 2007)

\bibitem{Wu2001}
Wu, O., Koroshetz, W.J., Ostergaard, L., Buonanno, F.S., Copen, W.a., Gonzalez,
  R.G., Rordorf, G., Rosen, B.R., Schwamm, L.H., Weisskoff, R.M., Sorensen,
  a.G.: {Predicting Tissue Outcome in Acute Human Cerebral Ischemia Using
  Combined Diffusion- and Perfusion-Weighted MR Imaging}. Stroke  32(4),
  933--942 (apr 2001)

\end{thebibliography}

\end{document}